\documentclass{article}
\usepackage{ijcai22}

\usepackage{times}

\usepackage{soul}
\usepackage{url}
\usepackage[hidelinks]{hyperref}
\usepackage[utf8]{inputenc}
\usepackage[small]{caption}
\usepackage{graphicx}
\usepackage{amsmath}
\usepackage{amsthm}

\usepackage{booktabs}
\usepackage{txfonts}
\usepackage{algorithm}
\usepackage{algorithmic}
\usepackage{makecell}
\usepackage{setspace}
\usepackage{threeparttable}
\usepackage{graphicx}
\usepackage{float}
\usepackage{subfigure}
\usepackage{url}
\newtheorem{thm}{Remark}
\title{Ensemble Multi-Relational Graph  Neural  Networks}

\author{
Yuling Wang$^{1,2}$
\and
Hao Xu$^2$\and
Yanhua Yu$^{1}$\footnote{Corresponding author.}\and
Mengdi Zhang$^2$\and
Zhenhao Li$^1$\and
Yuji Yang$^2$\\
\And Wei Wu $^2$ \\
\affiliations
$^1$Beijing University of Posts and Telecommunications\\
$^2$Meituan\\
\emails
wangyl0612@bupt.edu.com, 
kingsleyhsu1@gmail.com, 
yuyanhua@bupt.edu.com,
zhangmengdi02@meituan.com,
lzhbupt@bupt.edu.com,
\{yangyujiyyj, wuwei19850318\}@gmail.com
}

\begin{document}
\begin{sloppypar}

\maketitle

\begin{abstract}
It is well established that graph neural networks (GNNs) can be interpreted and designed from the perspective of optimization objective. With this clear optimization objective, the deduced GNNs architecture has sound theoretical foundation, which is able to flexibly remedy the weakness of GNNs. However, this optimization objective is only proved for GNNs with single-relational graph. \emph{Can we infer a new type of GNNs for multi-relational graphs by extending this optimization objective, so as to simultaneously solve the issues in previous multi-relational GNNs, e.g., over-parameterization? }
In this paper, we propose a novel ensemble multi-relational GNNs by designing an ensemble multi-relational (EMR)  optimization objective. 
This EMR optimization objective is able to derive an iterative updating rule, which can be formalized as an ensemble message passing (EnMP) layer with multi-relations. We further analyze the nice properties of EnMP layer, e.g., the relationship with multi-relational personalized PageRank. Finally, a new multi-relational GNNs which well alleviate the over-smoothing and over-parameterization issues are proposed.
Extensive experiments conducted on four benchmark datasets well demonstrate the effectiveness of the proposed model.\footnote{Code and appendix are at \url{https://github.com/tuzibupt/EMR}.}

\end{abstract}

\section{Introduction}
\label{sec:intro}
Graph neural networks (GNNs), which have been applied to a large range of downstream tasks, have displayed superior performance on dealing with graph data within recent years, e.g., biological networks~\cite{huang2020skipgnn} and knowledge graphs~\cite{yu2021knowledge}. Generally, the current GNN architecture follows the message passing frameworks, where the propagation process is the key component. For example, 
GCN~\cite{kipf2016semi} directly   aggregates and propagates transformed features along the topology at each layer.
PPNP~\cite{klicpera2018predict} aggregates both of the transformed features and the original features at each layer.
JKNet~\cite{xu2018representation}  selectively combines the aggregated messages from different layers via concatenation/max-pooling/attention operations.

Recent studies~\cite{zhu2021interpreting,ma2021unified} have proven that despite different propagation processes of various GNNs, they usually can be fundamentally unified as an optimization objective containing a  feature fitting term $O_{\text {fit }}$  and a graph regularization term $O_{\text {reg }} $ as follows:
\begin{eqnarray}
\boldsymbol{O}=\min _{\mathbf{Z}}\{\underbrace{\zeta\left\|\mathbf{F}_{1} \mathbf{Z}-\mathbf{F}_{2} \mathbf{H}\right\|_{F}^{2}}_{O_{\text {fit }}}+\underbrace{\xi \operatorname{tr}\left(\mathbf{Z}^{T} \tilde{\mathbf{L}} \mathbf{Z}\right)}_{O_{\text {reg }}}\},
\end{eqnarray}
where $\mathbf{H}$ is the original input feature and $\mathbf{L}$ is the graph Laplacian matrix encoding the graph structure. $\mathbf{Z}$ is the propagated representation, and $\mathbf{F}_{1}$, $\mathbf{F}_{2}$ are defined as arbitrary graph convolutional kernels and usually set as $\mathbf{I}$.
This optimization objective reveals a mathematical guideline that essentially governs the propagation mechanism, and opens a new path to design novel GNNs. That is, such clear optimization objective is able to derive the corresponding propagation process, further making the designed GNN architecture more interpretable and reliable~\cite{zhu2021interpreting,liu2021elastic,yang2021graph}.
For example,~\cite{zhu2021interpreting} replaces $\mathbf{F}_1$ and $\mathbf{F}_2$ with high-pass kernel and infers new high-pass GNNs; ~\cite{liu2021elastic} applies $l_{1}$ norm to  $O_{\text {reg }} $ term and infers Elastic GNNs. 

Despite the great potential of this optimization objective on designing GNNs, it is well recognized that it is only proposed for traditional homogeneous graphs, rather than the multi-relational graphs with multiple types of relations. 
However, in real-world applications, multi-relational graphs  tend to be more general and pervasive in many areas. 
For instance, the various types of chemical bonds in molecular graphs,
and the diverse relationships between people in social networks.
Therefore, it is greatly desired to design GNN models that are able to adapt to multi-relational graphs.
Some literatures have been devoted to the multi-relational GNNs, which can be roughly categorized into  feature mapping based approaches~\cite{schlichtkrull2018modeling} and learning relation embeddings based approaches~\cite{vashishth2019composition}.
However, these methods usually design the propagation process heuristically without a clear and  an explicit mathematical objective. Despite they improve the performance, they still suffer from the problems of over-parameterization~\cite{vashishth2019composition} and over-smoothing~\cite{oono2019graph}.

\emph{``Can we remedy the original optimization objective to design a new type of multi-relational GNNs that is more reliable with solid objective, and at the same time, alleviates the weakness of current multi-relational GNNs, e.g., over-smoothing and over-parameterization?"}

Nevertheless, it is technically challenging to achieve this goal. Firstly, how to incorporate multiple relations into an optimization objective. Different relations play different roles, and we need to distinguish them in this optimization objective as well. Secondly, to satisfy the above requirements, it is inevitable that the optimization objective will become more complex, maybe with more constrains. How to derive the underlying message passing mechanism by optimizing the objective is another challenge. Thirdly, even with the message passing mechanism, 
it is highly desired that how to integrate it into deep neural networks via simple operations without introducing excessive parameters.

In this paper, we propose a novel multi-relational GNNs by designing an ensemble optimization objective. In particular, our proposed ensemble optimization objective consists of a feature fitting term and an ensemble multi-relational graph regularization (EMR) term. Then we derive an iterative optimization algorithm with this ensemble optimization objective to learn 
the node representation and the relational coefficients as well. We further show that this iterative optimization algorithm can be formalized as an ensemble message passing layer, which has a nice relationship with multi-relational personalized PageRank and covers some existing propagation processes. Finally, we integrate the derived ensemble message passing layer into deep neural networks by decoupling the feature transformation and  message passing process, and
a novel family of multi-relational GNN architectures is  developed.
Our key contributions can be summarized as follows:

\begin{itemize}
    \item 
    We make the first effort on how to derive multi-relational GNNs from the perspective of optimization framework, so as to enable the derived multi-relational GNNs more reliable. This research holds great potential for opening new path to design multi-relational GNNs.
    \item

    We propose a new optimization objective for multi-relational graphs, and we derive a novel ensemble message passing (EnMP) layer. A new family of multi-relational GNNs is then proposed in a decoupled way.
    
    \item
    We build the relationships between our proposed EnMP layer with multi-relational personalized PageRank, and some current message passing layers. Moreover, our proposed multi-relational GNNs can well alleviate the over-smoothing and over-parameterazion issues.
    
     \item 

   Extensive experiments are conducted, which comprehensively demonstrate the effectiveness of our proposed multi-relational GNNs.
 
\end{itemize}

\section{Related Work}
\label{sec:related}
\noindent
\textbf{Graph Neural Networks.}
The dominant paradigms of GNNs can be generally summarized into two branches: spectral-based GNNs~\cite{defferrard2016convolutional,klicpera2018predict} and spatial-based GNNs~\cite{gilmer2017neural,klicpera2018predict}. 
Various of representative GNNs have been proposed by designing different information aggregation and update strategies along  topologies, e.g.,~\cite{gilmer2017neural,klicpera2018predict}.
Recent works~\cite{zhu2021interpreting,ma2021unified}  have explore the intrinsically unified  optimization framework behind existing GNNs. 

\noindent
\textbf{Multi-relational Graph Neural Networks.}
The core idea of multi-relational GNNs~\cite{schlichtkrull2018modeling,vashishth2019composition,thanapalasingam2021relational} is to encode relational graph structure information into low-dimensional node or relation embeddings.
As a representative relational GNNs, RGCN~\cite{schlichtkrull2018modeling} designs a specific convolution for each relation, and then the convolution results under all relations are aggregated, these excess parameters generated are completely learned in an end-to-end manner.
Another line of literature~\cite{ji2021heterogeneous,wang2019heterogeneous,fu2020magnn,yun2019graph} considers the heterogeneity of edges and nodes to construct meta-paths, then aggregate messages from different meta-path based  neighbors.

\section{Proposed Method}
\label{sec:proposed}
\textbf{Notations.} Consider a  multi-relational graph $\mathbf{\mathcal{G}}=(\mathcal{V}, \mathcal{E}, \mathcal{R})$ with nodes $v_{i} \in \mathcal{V}$ and labeled edges (relations) $\left(v_{i}, r, v_{j}\right) \in \mathcal{E}$ , where $r \in \mathcal{R}$ is a relation type. 
Graph structure $\mathcal{G}^{r}$ under relation $r$ can be described by the adjacency matrix $\mathbf{A}^{r} \in \mathbb{R}^{n \times n}$, where  $\mathbf{A}^r_{{i, j}}=1$ if there is an edge between nodes $i$ and $j$ under relation $r$, otherwise $0$. 
The diagonal degree matrix is denoted as $\mathbf{D}^{r}=\operatorname{diag}\left(d^{r}_{1}, \cdots, d^{r}_{n}\right)$, 
where $d^{r}_{j}=\sum_{j} \mathbf{A}^{r}_{i, j}$.
We use $\tilde{\mathbf{A}}^{r}=\mathbf{A}^{r}+\mathbf{I}$ to represent the adjacency matrix with added self-loop and $\tilde{\mathbf{D}}^{r}=\mathbf{D}^{r}+\mathbf{I}$. 
Then the normalized adjacency matrix is $\hat{\tilde{\mathbf{A}}}^{r}=\left(\tilde{\mathbf{D}}^{r}\right)^{-1 / 2} \tilde{\mathbf{A}}^{r} \left(\tilde{\mathbf{D}}^{r}\right)^{-1 / 2}$. Correspondingly, $\tilde{\mathbf{L}}^{r}=\mathbf{I}-\hat{\tilde{\mathbf{A}}}^{r}$ is the normalized symmetric positive semi-definite graph Laplacian matrix of relation $r$.

\subsection{Ensemble Optimization Framework}

Given a multi-relational graph, one basic requirement is that the learned representation $\mathbf{Z}$ should capture the homophily property in the graph with relation $r$, i.e., the representations $\mathbf{Z}_i$ and $\mathbf{Z}_j$ should be similar if nodes $i$ and $j$ are connected by relation $r$. We can achieve the above goal by minimizing to the following term with respect to $\mathbf{Z}$:
\begin{eqnarray}
\operatorname{tr}\left(\mathbf{Z}^{T} \tilde{\mathbf{L}}^r \mathbf{Z}\right)=\sum_{i, j}^{n} \hat{\tilde{\mathbf{A}}}_{i, j}^r\left\|\mathbf{Z}_{i}-\mathbf{Z}_{j}\right\|^{2},
\end{eqnarray}
where $\hat{\tilde{\mathbf{A}}}_{i, j}^r$ represents the node $i$ and node $j$ are connected under relation $r$.

With all the $R$ relations, we need to simultaneously capture the graph signal smoothness. Moreover, consider that different relations may play different roles, we need to distinguish their importance as well, which can be modelled as an ensemble multi-relational graph  regularization as follows: 
\begin{eqnarray}
\mathcal{O}_{\text {e-reg }} = 
\left\{
\begin{aligned}
 \lambda_{1} \sum_{r=1}^{R} \mu_{r}\sum_{i, j}^{n} \hat{\tilde{\mathbf{A}}}_{i, j}^{r}\left\|\mathbf{Z}_{i}-\mathbf{Z}_{j}\right\|^{2}
+\lambda_{2}\left\|\boldsymbol{\mu}\right\|_{2}^{2},\\
\text { s.t. } \sum_{r=1}^{R}\mu_{r}=1, \mu_{r} \geq 0, \forall r=1,2, \ldots, R,
\end{aligned}
\right.
\end{eqnarray}
where $\mathbf{\lambda}_{1}$ and $\mathbf{\lambda}_{2}$ are  non-negative trade-off parameters.
$\mu_{r} \geq 0$ is the weight corresponding to relation $r$, and 
the sum of weights is 1 for  constraining the search space of possible graph Laplacians.
The regularization term $\left\|\boldsymbol{\mu}\right\|_{2}^{2}$ is to avoid the parameter overfitting to only one relation~\cite{geng2012ensemble}.


In addition to the topological constraint by $\mathcal{O}_{\text {e-reg }}$ term, we should also build the relationship between the learned representation $\mathbf{Z}$ with the node features $\mathbf{H}$. Therefore, there is a feature fitting term: $\mathcal{O}_{\text {fit }} =\|\mathbf{Z-H}\|_{F}^{2}$,
which makes $\mathbf{Z}$  encode information from the original feature $\mathbf{H}$, so as to  alleviate the over-smoothing problem.
Finally, our proposed optimization framework for multi-relational graphs, which  includes constraints on features and  topology, is as follows:
\begin{eqnarray}
\label{overall}
\notag& \arg&\min_{\mathbf{Z},\mathbf{\mu}}\underbrace{\|\mathbf{Z-H}\|_{F}^{2}}_{\mathcal{O}_{\text {fit }}}+
\underbrace{\lambda_{1} \sum_{r=1}^{R} \mu_{r} \mathbf{tr}\left(\mathbf{Z}^{\top}\tilde{\mathbf{L}}^{r} \mathbf{Z}\right)+\lambda_{2}\left\|\boldsymbol{\mu}\right\|_{2}^{2}}_{\mathcal{O}_{\text {e-reg }}}, \\
 &\text { s.t. }& \sum_{r=1}^{R}\mu_{r}=1, \mu_{r} \geq 0, \forall r=1,2, \ldots, R.
\end{eqnarray}

By minimizing the above objective function,  the optimal representation $\mathbf{Z}$ not only captures the smoothness between nodes, but also preserves the personalized information. Moreover, the optimal relational coefficients $\boldsymbol{\mu}$ can be derived, reflecting the importance of different relations.

\subsection{Ensemble Message Passing Mechanism }

It is nontrivial to directly optimize $\mathbf{Z}$ and $\boldsymbol{\mu}$ together because Eq.\eqref{overall} is not convex w.r.t. $ (\mathbf{Z}, \boldsymbol{\mu})$ jointly.
Fortunately, an iterative optimization strategy can be adopted,
i.e., i.) first optimizing Eq.\eqref{overall} w.r.t. $ \boldsymbol{\mu}$ with a fixed  $\mathbf{Z}$,
resulting in the solution of relational coefficients $ \boldsymbol{\mu}$;
ii.) then solving Eq.\eqref{overall} w.r.t. $\mathbf{Z}$  with $ \boldsymbol{\mu}$ taking the value solved in the last iteration.
We will show that performing the above two steps corresponds to one ensemble message  passing layer in our relational GNNs.


\setcounter{secnumdepth}{3}
\subsubsection*{Update Relational Coefficients}



We update relational parameters $ \boldsymbol{\mu}$ by fixing $\mathbf{Z}$, then the objective function \eqref{overall} w.r.t. $ \boldsymbol{\mu}$  is reduced to:
\begin{eqnarray}
\label{cor} 
\notag &\arg& \min _{\mu} \sum_{r=1}^{R} \mu_{r} s_{r} +
\mathbf{\frac{\lambda_{2}}{\lambda_{1}}}\left\|\boldsymbol{\mu}\right\|_{2}^{2},\\
 &\text { s.t. }& \sum_{r=1}^{R}\mu_{r}=1, \boldsymbol{\mu} \geq 0, \forall r=1,2, \ldots, R,
\end{eqnarray}
where $s_{r}=\mathbf{tr}\left(\mathbf{Z}^{\top}\tilde{\mathbf{L}}^{r} \mathbf{Z}\right)$.

(1) When $\mathbf{\frac{\lambda_{2}}{\lambda_{1}}}=0$, 
the coefficient might concentrate on one certain relation,
i.e.,  $\mu_{j}=1$ if ${s}_{j}=\min_{r=1, \ldots, R} {s}_{r}$, and $\mu_{j}=0$ otherwise. 
When $\mathbf{\frac{\lambda_{2}}{\lambda_{1}}}=+\infty$, each relation will be assigned equal coefficient, i.e., $\mu_r=\frac{1}{R}$~\cite{geng2012ensemble}.

(2) Otherwise, theoretically, Eq.\eqref{cor} can be regarded as a  convex function of $\boldsymbol{\mu}$ with the constraint in a standard simplex~\cite{chen2011projection},  i.e., $\Delta=$ $\left\{\boldsymbol{\mu} \in \mathbb{R}^{R}: \sum_{r=1}^{R} \mu_{r}=1, \boldsymbol{\mu} \succcurlyeq 0\right\}$.
Therefore, the mirror entropic descent algorithm (EMDA)~\cite{beck2003mirror} can be used to optimize $\boldsymbol{\mu}$, where the update process is described by Algorithm~\ref{alg1}.
The objective $f(\cdot)$ should be a convex Lipschitz continuous function with Lipschitz constant $\phi$ for a fixed given norm. Here, we derive this Lipschitz constant from $\|\nabla f(\boldsymbol{\mu})\|_{1} \leq  \frac{2\lambda_{2}}{\lambda_{1}}+\|\boldsymbol{s}\|_{1}=\phi$, where $\boldsymbol{s}=\left\{s_{1}, \ldots, s_{R}\right\}$.

\subsubsection*{Update Node Representation}
Then we update node representation $\mathbf{Z}$ with  fixing $\boldsymbol{\mu}$, where
the objective function Eq. \eqref{overall} w.r.t.  $\mathbf{Z}$ is reduced to:
\begin{eqnarray}
\label{z}
\arg \min _{\mathbf{Z} }\|\mathbf{Z}-\mathbf{H}\|_{F}^{2}+
\lambda_{1} \sum_{r=1}^{R} \mu_{r} \mathbf{tr}\left(\mathbf{Z}^{\top}\tilde{\mathbf{L}}^{r} \mathbf{Z}\right)
.\end{eqnarray}
We can set the derivative of Eq.~\eqref{z} with respect to $\mathbf{Z}$ to zero and get the optimal $\mathbf{Z}$ as:

\begin{eqnarray}
\frac{\partial\left\{
\|\mathbf{Z}-\mathbf{H}\|_{F}^{2}+
\lambda_{1} \sum_{r=1}^{R} \mu_{r} \mathbf{tr}\left(\mathbf{Z}^{\top}\tilde{\mathbf{L}}^{r} \mathbf{Z}\right)
\right\}}{\partial \mathbf{Z}}=0\\ \quad \Rightarrow \quad \mathbf{Z}-\mathbf{H} + \lambda_{1}\sum_{r=1}^{R} \mu_{r}\tilde{\mathbf{L}}^{r} \mathbf{Z}=0.
\end{eqnarray}
Since the eigenvalue of $\mathbf{I}+\lambda_{1}\sum_{r=1}^{R} \mu_{r}\tilde{\mathbf{L}}^{r}$ is positive, it has an inverse matrix, and we can obtain the closed solution as follows:
\begin{eqnarray}
\label{close}
\notag\mathbf{Z}&=&\left(\mathbf{I}+\lambda_{1}  \sum_{r=1}^{R} \mu_{r}\tilde{\mathbf{L}}^{r} \right)^{-1} \mathbf{H}\\ &=&\frac{1}{1+\lambda_{1}}\left(\mathbf{I}-\frac{\lambda_{1}}{1+\lambda_{1}} \sum_{r=1}^{R} \mu_{r}\hat{\tilde{\mathbf{A}}}^{r}\right)^{-1} \mathbf{H}.
\end{eqnarray}

However, obtaining the inverse of matrix will cause  a
computational complexity and memory requirement of  $\mathcal{O}\left(n^{2}\right)$, 
which is inoperable in large-scale graphs. 
Therefore,  we can approximate Eq.\eqref{close} using the following iterative update rule:
\begin{eqnarray}
\label{update}
\mathbf{Z}^{(k+1)}=\frac{1}{\left(1+\lambda_{1}\right)}\mathbf{H}+\frac{\lambda_{1}}{\left(1+\lambda_{1}\right)}\left(\sum_{r=1}^{R} \mu_{r}^{(k)}\hat{\tilde{\mathbf{A}}}^{r}\right)\mathbf{Z}^{(k)}.
\end{eqnarray}
where $k$ is the iteration number.

\begin{algorithm}
	\renewcommand{\algorithmicrequire}{\textbf{Input:}}
	\renewcommand{\algorithmicensure}{\textbf{Output:}}
	\caption{Relational Coefficients Learning}
	\label{alg1}
	\begin{algorithmic}[1]
	    \REQUIRE 
	    Candidate Laplacians $\left\{\tilde{\mathbf{L}}^{1},\cdots, \tilde{\mathbf{L}}^{R} \right\}$, 
	    the embedding matrix $\mathbf{Z}$,
	    the Lipschitz constant $\phi$, the tradeoff parameters $\lambda_{1},\lambda_{2}$.
	    \ENSURE  Relational coefficients $\boldsymbol{\mu}$.
		\STATE Initialization: $\boldsymbol{\mu}=[\frac{1}{R},\frac{1}{R},\cdots,\frac{1}{R}] $
		\FOR { $r=1$ to $R$}
		\STATE $s_{r}=\mathbf{tr}\left(\mathbf{Z}^{\top}\tilde{\mathbf{L}}^{r} \mathbf{Z}\right)$
                \REPEAT
                \STATE  $T_{t} \leftarrow \sqrt{\frac{2 ln R}{t \phi^{2}}}$,
            \STATE  $f^{\prime}\left(\mu_{r}^{t}\right) \leftarrow \frac{2\lambda_{2}}{\lambda_{1}} \mu_{r}^{t}+s_{r}$,
            \STATE   $\mu_{r}^{t+1} \leftarrow \frac{\mu_{r}^{t} \exp \left[-T_{t} f^{\prime}\left(\mu_{r}^{t}\right)\right]}{\sum_{r=1}^{R} \mu_{r}^{t} \exp \left[-T_{t} f^{\prime}\left(\mu_{r}^{t}\right)\right]}$,
            \UNTIL Convergence
        \ENDFOR
        \RETURN $\boldsymbol{\mu}$
	\end{algorithmic}  
\end{algorithm}

\subsubsection*{Ensemble Message Passing Layer (EnMP layer)}
Now with the node representation $\mathbf{Z}$ and the relation coefficient $\boldsymbol{\mu}$, we can propose our ensemble message passing layer, consisting of the following two steps: (1) relational coefficient learning step (RCL step), i.e., update the relational coefficients $\boldsymbol{\mu}$ according to Algorithm~\ref{alg1}; (2) propagation step (Pro step), i.e., update the node representation $\mathbf{Z}$ according to Eq.\eqref{update}. The pseudocode of EnMP layer is shown in appendix A. We will show some properties of our proposed EnMP layer. 


\begin{thm}[Relationship with Multi-Relational/Path Personalized PageRank]
\label{reppr}
Given a realtion $r$, we have the relation based probability transition  matrix $\mathbf{A}^{r}_{rw}=\mathbf{A}^{r}(\mathbf{D}^{r})^{-1} $. 
Then, the single relation based PageRank
matrix is calculated via:
\begin{eqnarray}
\boldsymbol{\Pi}_{\mathrm{pr}}^{r}=\mathbf{A}^{r}_{\mathrm{rw}}\boldsymbol{\Pi}_{\mathrm{pr}}^{r}.
\end{eqnarray}
Considering we have $R$ relations, i.e., $r=1,2, \ldots, R$, the weights of each relation are $\left\{\mu_{1}, \ldots, \mu_{R}\right\}$, according to~\cite{lee2013pathrank,ji2021heterogeneous}, we can define the multiple relations based PageRank matrix:
\begin{eqnarray}
\boldsymbol{\Pi}_{\mathrm{pr}}=\left(\sum_{r=1}^{R} \mu_{r}\mathbf{A}^{r}_{\mathrm{rw}}\right)
\boldsymbol{\Pi}_{\mathrm{pr}}^{r}.
\end{eqnarray}
According to~\cite{klicpera2018predict}, the  multi-relational personalized PageRank matrix can be defined:
\begin{eqnarray}
\boldsymbol{\Pi}_{\mathrm{ppr}}=\alpha\left(\boldsymbol{I}_{n}-(1-\alpha)
\left(\sum_{r=1}^{R} \mu_{r}\hat{\tilde{\boldsymbol{A}}}^{r}\right)
\right)^{-1},
\end{eqnarray}
where $\hat{\tilde{\boldsymbol{A}}}^{r}$ is a normalized adjacency matrix with self-loops, $\boldsymbol{I}_{n}$ represents unit matrix, $\alpha \in(0,1]$ is teleport (or restart) probability.
If $\alpha=\frac{1}{\left(1+\lambda_{1}\right)}$,  
the closed-form solution in Eq.\eqref{close} is to propagate features via multi-relational personalized PageRank scheme.

\end{thm}
\begin{thm}[Relationship with APPNP/GCN]
if ${\lambda_{2}}=+\infty$, the solution in Eq.\eqref{cor} is $\boldsymbol{\mu}=[\frac{1}{R},\frac{1}{R},\cdots,\frac{1}{R}] $,
i.e., each relaiton  is assigned equal coefficient,
then the ensemble multi-relational graph $\sum_{r=1}^{R} \mu_{r}^{(k)}\hat{\tilde{\mathbf{A}}}^{r}$ reduces to a normalized adjacency matrix $ \frac{1}{R} \sum_{r=1}^{R} \hat{\tilde{\mathbf{A}}}^{r}$  averaged over all relations. The proposed message passing scheme reduces to:
\begin{eqnarray}
\mathbf{Z}^{(k+1)}=\frac{1}{\left(1+\lambda_{1}\right)}\mathbf{H}+\frac{\lambda_{1}}{\left(1+\lambda_{1}\right)}\frac{1}{R} \sum_{r=1}^{R} \hat{\tilde{\mathbf{A}}}^{r}\mathbf{Z}^{(k)}
,\end{eqnarray}
if $\lambda_{1}=\frac{1}{\alpha}-1$,  it recovers the message passing in APPNP on the averaged relational graph:
\begin{eqnarray}
\mathbf{Z}^{(k+1)}=\alpha\mathbf{H}+(1-\alpha)\frac{1}{R} \sum_{r=1}^{R} \hat{\tilde{\mathbf{A}}}^{r}\mathbf{Z}^{(k)},
\end{eqnarray}
if $\lambda_{1}= +\infty $,  it recovers the message passing in GCN on the averaged relational graph:
\begin{eqnarray}
\mathbf{Z}^{(k+1)}=\frac{1}{R} \sum_{r=1}^{R} \hat{\tilde{\mathbf{A}}}^{r}\mathbf{Z}^{(k)}.
\end{eqnarray}

\end{thm}

\begin{figure}[htp]
\centering
    \includegraphics[width = 8cm,height = 2.8cm]{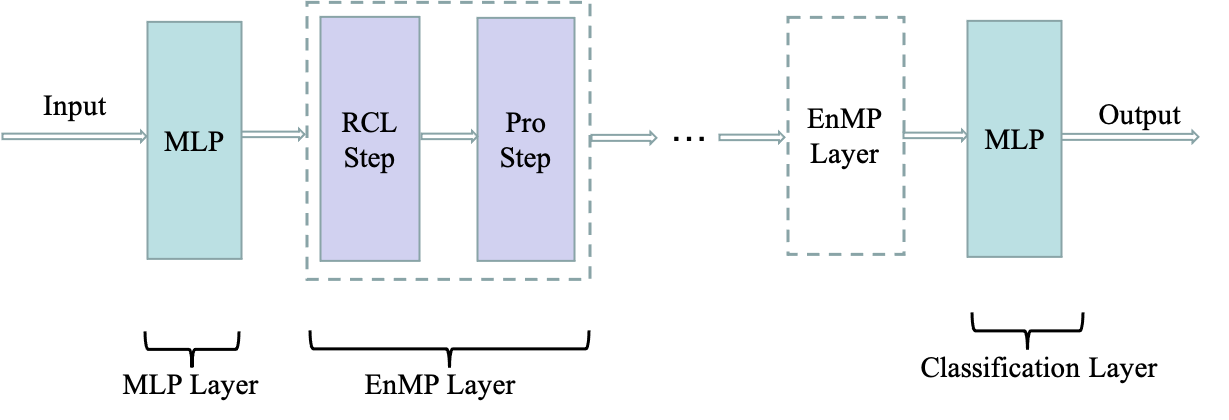}
    \caption{Model architecture.}
    \label{model}
\end{figure}

\subsection{Ensemble Multi-Relational GNNs}

Now we propose our ensemble multi-relational graph neural networks (EMR-GNN) with the EnMP layer. Similar as \cite{klicpera2018predict}, we employ the decoupled style architecture, i.e., the feature transformation and the message passing layer are separated. The overall framework is shown in Figure~\ref{model}, and the forward propagation process is as follows:
\begin{eqnarray}
\mathbf{Y}_{\text {pre }}= g_{\theta}\left(
\mathbf{EnMP}^{(K)}\left(f\left(\mathbf{X};\mathbf{W}\right), R, \lambda_{1}, \lambda_{2}\right) \right),
\end{eqnarray}
where $\mathbf{X}$ is the input feature of nodes,
and $f\left(\mathbf{X};\mathbf{W}\right)$ denotes the MLPs or linear layers (parameterized by $ \mathbf{W}$)  which is used to feature extraction.
$\mathbf{EnMP}^{(K)}$ represents our designed ensemble relational message passing layer with $K$ layers, where $R$ is the number of relations, and $\lambda_{1}, \lambda_{2}$ are hyperparameters in our message passing layer.
$g_{\theta}(\cdot)$ is MLPs as classifier with the learnable parameters $\theta$.
The training loss is: $\ell(\mathbf{W}, \theta) \triangleq  \mathcal{D}\left(\boldsymbol{y}^{*}_{i}, \hat{\boldsymbol{y}_{i}}\right),$
where $\mathcal{D}$ is a discriminator function of cross-entropy, 
$\boldsymbol{y}^{*}_{i}$ and $ \hat{\boldsymbol{y}_{i}}$ are the predicted and ground-truth labels of node $i$, respectively.
Backpropagation manner is used to optimize parameters in MLPs, i.e., $\mathbf{W}$ and $\theta$,
and the parameters in our EnMP layers are optimized during the forward propagation. We can see that EMR-GNN is built on a clear optimization objective. Besides, EMR-GNN also has the following two advantages:
\begin{itemize}
    
    \item  As analyzed by Remark~\ref{reppr}, our proposed EnMP can  keep  the  original  information  of  the  nodes with  a  teleport (or restart) probability,  thereby  alleviating over-smoothing.
    \item For each relation, there is a parameterized relation-specific weight matrix or parameterized relation encoder used in the traditional RGCN~\cite{vashishth2019composition,schlichtkrull2018modeling}. While in our EnMP, only one learnable weight coefficient is associated with a relation, greatly alleviating the over-parameterization problem.

\end{itemize}

\begin{table}
   \centering
    \renewcommand{\arraystretch}{1.2}
    \small 
    \resizebox{8cm}{!}{
    \begin{tabular}{lrrrrlr}
        \toprule
        Datasets &Nodes &\makecell*[r]{Node\\Types}& Edges&\makecell*[c]{ Edge \\Types} & Target & Classes\\
        \midrule
        MUTAG & 23,644 & 1 & 74,227&23&molecule &2\\
        BGS & 333,845& 1 & 916,199&103&rock &2\\
        DBLP&26,128&4&239,566&6&author&4\\
        ACM&10,942&4&547,872&8&paper&3\\
        \bottomrule
    \end{tabular}}
    \caption{Statistics of multi-relational datasets.}
    \label{sta}
\end{table}

\begin{table*}[!htb]
\centering
\renewcommand{\arraystretch}{1.2}
\setlength\tabcolsep{2.6pt}

\small
\begin{tabular}{l|r|r|r|r|r|r|r|r}
\bottomrule 
Dataset &\multicolumn{2}{c|}{DBLP} &\multicolumn{2}{c|}{ACM}&\multicolumn{2}{c|}{MUTAG}&\multicolumn{2}{c}{BGS}\\
\hline
Metric &{Acc ($\%$)}& Recall ($\%$) &Acc ($\%$)& Recall ($\%$)&Acc ($\%$)& Recall ($\%$)&Acc ($\%$)&Recall ($\%$)\\
\hline
      GCN & 90.39±0.38 & 89.49±0.52  &89.58±1.47 &89.47±1.49 &72.35±2.17 &63.28±2.95 &85.86±1.96&80.21±2.21\\
     
  GAT & 91.97±0.40 & 91.25±0.58 &88.99±1.58 &88.89±1.56 & 70.74±2.13&63.01±3.79 &88.97±3.17&86.13±4.96\\
  HAN &91.73±0.61  &91.15±0.72  & 88.51±0.35 &88.50±0.30 &- &- &-&-\\
  RGCN & 90.08±0.60 &88.56±0.76& 89.79±0.62&89.71±0.59 &71.32±2.11 & 61.97±3.52&85.17±5.87& 81.58±7.94\\

  e-RGCN & 91.77±0.90& 91.18±1.02 &83.00±1.04 &84.03±0.75 &69.41±2.84& \textbf{67.57±8.04} &82.41±1.96&84.51±3.38\\

   \hline
 EMR-GNN &  \textbf{93.54±0.50} & \textbf{92.39±0.78  } & \textbf{90.87±0.11} &  \textbf{90.84±0.13} & \textbf{74.26±0.78} & 64.19±1.08&\textbf{89.31±4.12}&\textbf{86.39± 5.33} \\
\bottomrule 
\end{tabular}
\caption{ 
The mean and standard deviation of classification
accuracy and recall over 10 different runs on four datasets.}
\label{total}
\end{table*}

\begin{table*}[!htb]
\centering
\renewcommand{\arraystretch}{1.2}

\small 

\begin{tabular}{cccccc}

\bottomrule 
 GCN&GAT& HAN &RGCN & e-RGCN &EMR-GNN\\
\hline
$\mathcal{O}\left(K d^{2}\right)$&
$\mathcal{O}\left(2 K N d^{2}\right)$&
$\mathcal{O}\left(
2K\left(|\mathcal{R}| N +1 \right)d^{2} + 2 Kd
\right)$&
$\mathcal{O}\left(\mathcal{B} K d^{2}+\mathcal{B} K|\mathcal{R}|\right)$&
$\mathcal{O}\left(\mathcal{B} (K-1) d^{2}+|\mathcal{R}|d+\mathcal{B} (K-1)|\mathcal{R}|\right)$&
$\mathcal{O}\left(2 d^{2}+K|\mathcal{R}|\right)$

\\

\bottomrule 
\end{tabular}
\caption{
Comparison of the number of parameters.
Here, $K$ denotes the number of layers in the model, $d$ is the embedding dimension, $\mathcal{B}$ represents the number of bases, $|\mathcal{R}|$ indicates the total number of relations in the graph and $N$ is the number of heads of attention-based models.}
\label{pa}
\end{table*}

\begin{figure*}[t]
\centering  
\subfigure[GCN (SC=0.6196)]{
\label{Fig.sub.2}
\includegraphics[width=4cm,height = 3cm]{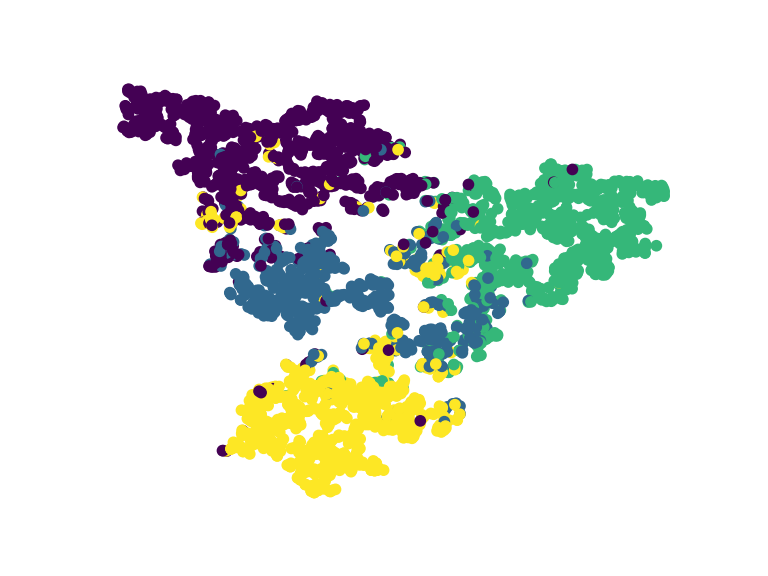}}
\subfigure[GAT (SC=0.6351)]{
\label{Fig.sub.2}
\includegraphics[width=4cm,height = 3cm]{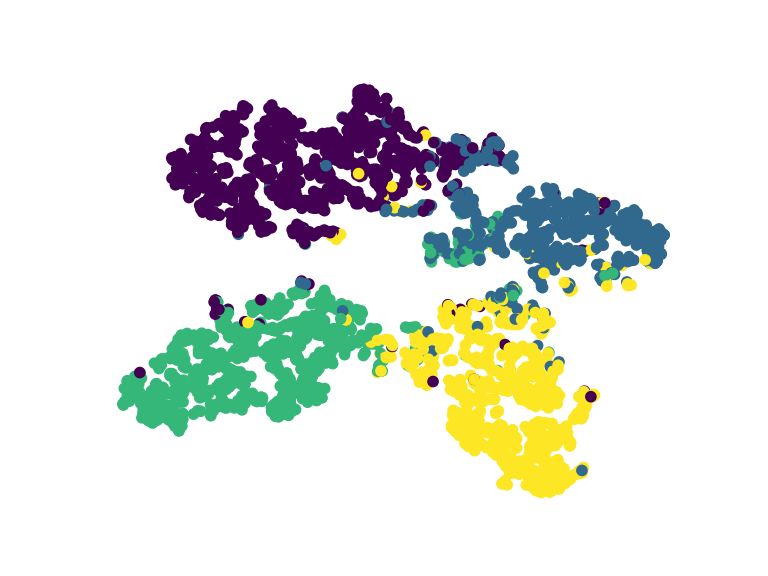}}
\subfigure[RGCN (SC=0.6093)]{
\label{Fig.sub.2}
\includegraphics[width=4cm,height = 3cm]{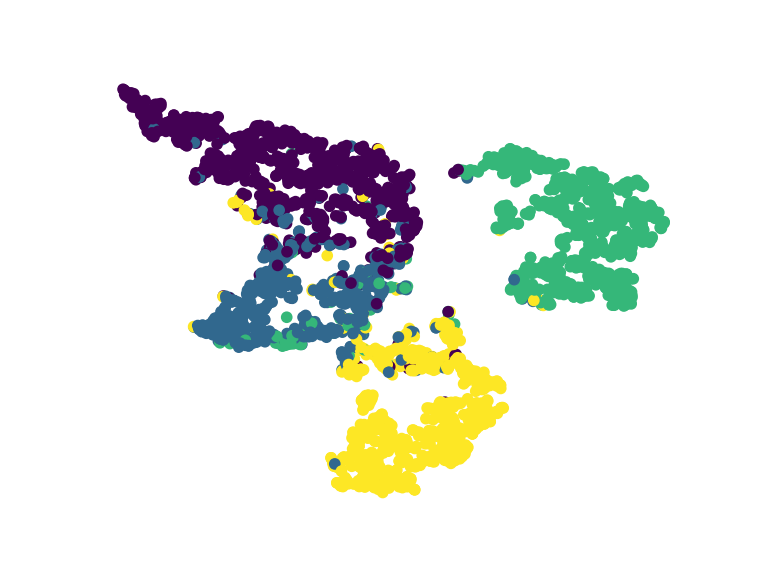}}
\subfigure[EMR-GNN (SC=0.7799)]{
\label{Fig.sub.1}
\includegraphics[width=4cm,height = 3cm]{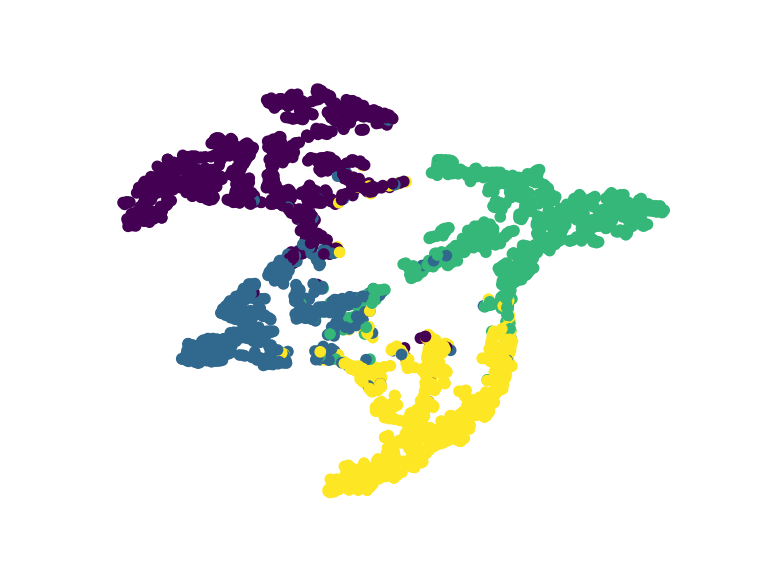}}
\caption{Visualization of the learned node embeddings on DBLP dataset.}
\label{vis}
\end{figure*}

\begin{figure}[t]
\centering  
\subfigure[DBLP]{
\label{Fig.sub.1}
\includegraphics[width=4.15cm,height = 4.cm]{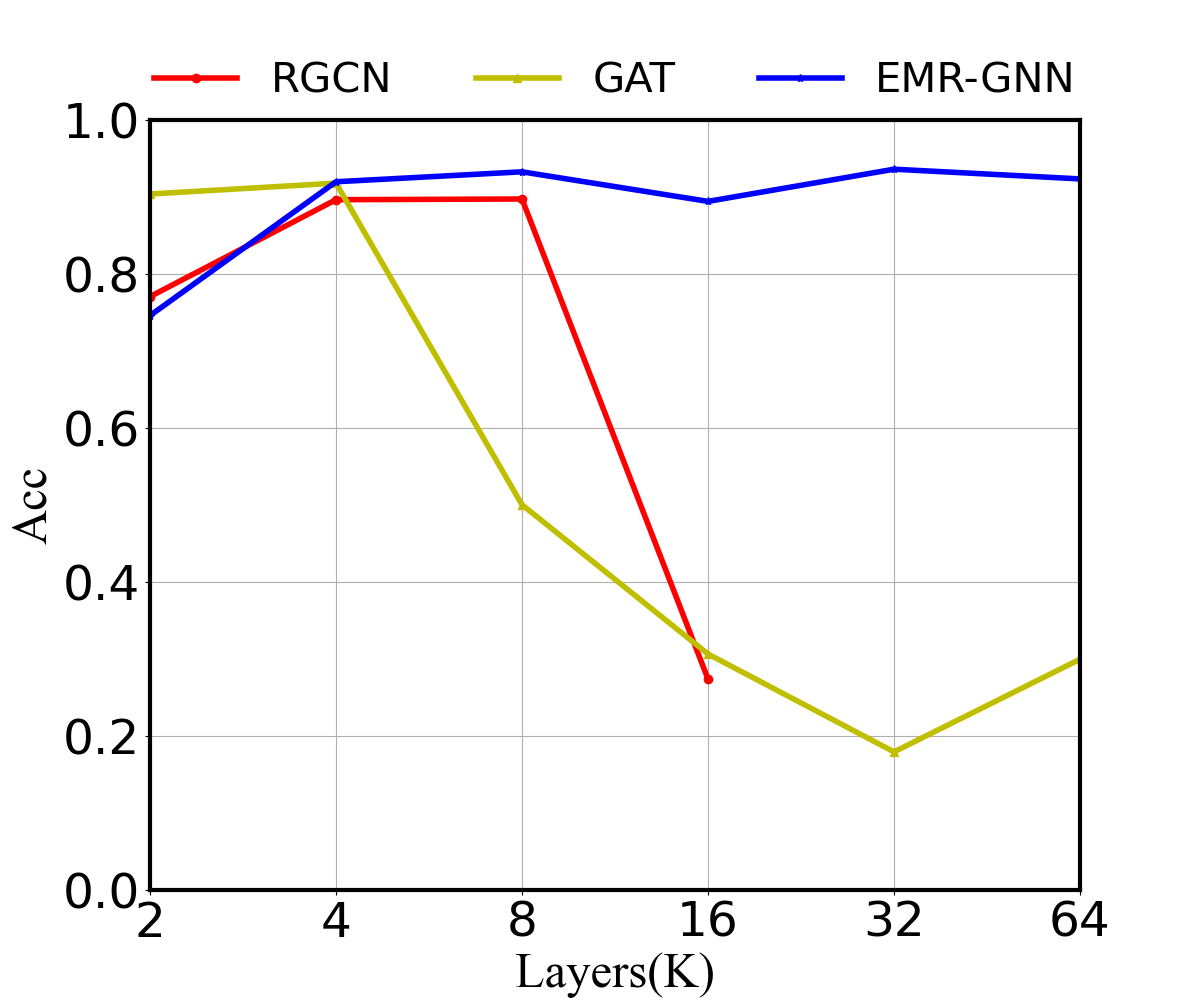}}
\subfigure[ACM]{
\label{Fig.sub.2}
\includegraphics[width=4.15cm,height = 4.cm]{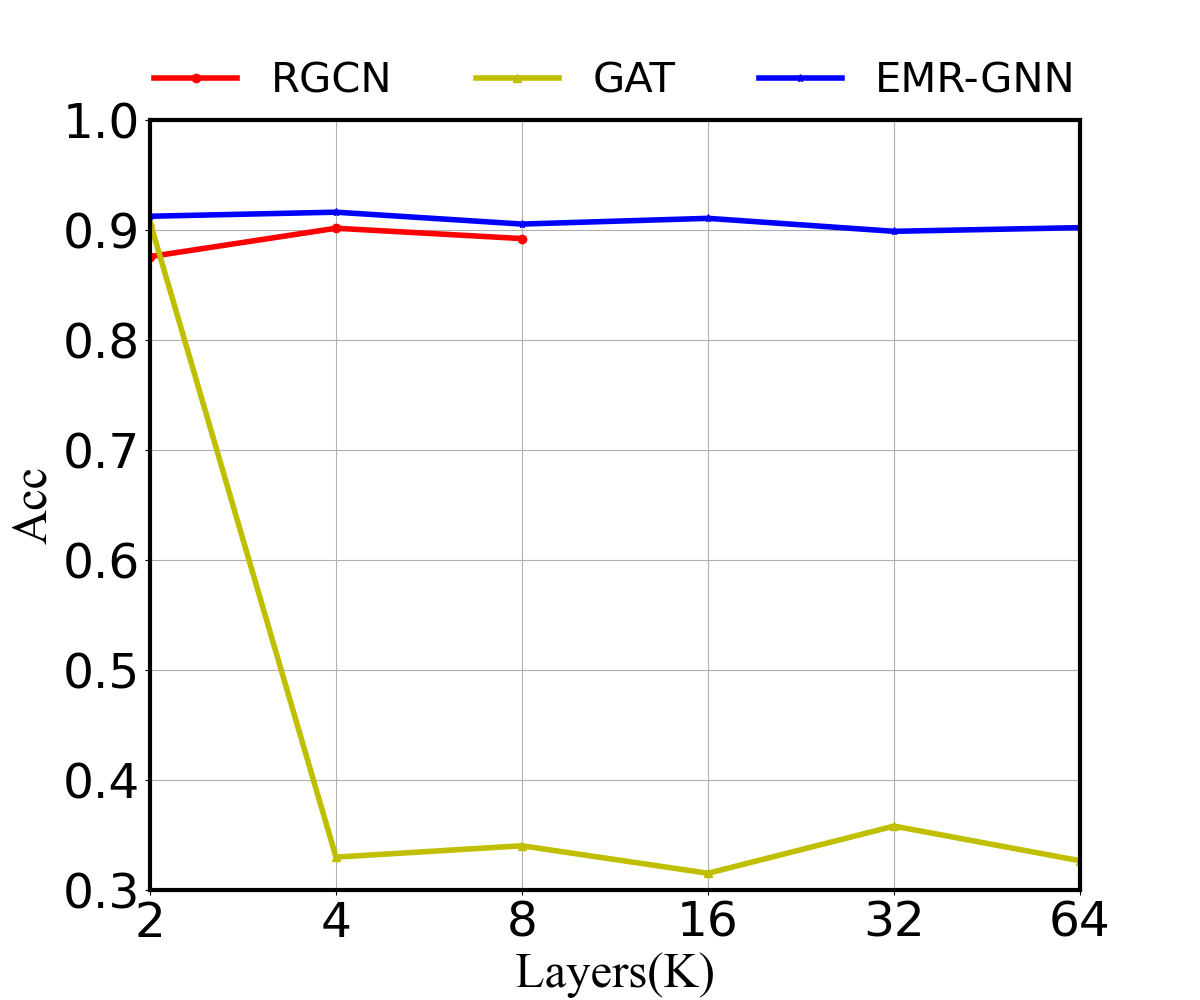}}
\caption{ Analysis of propagation layers. Missing value in red line means CUDA is out of memory.}
\label{nlayer}
\end{figure}

\begin{figure}[htp]
    \centering
    \includegraphics[width=7cm,height = 5cm]{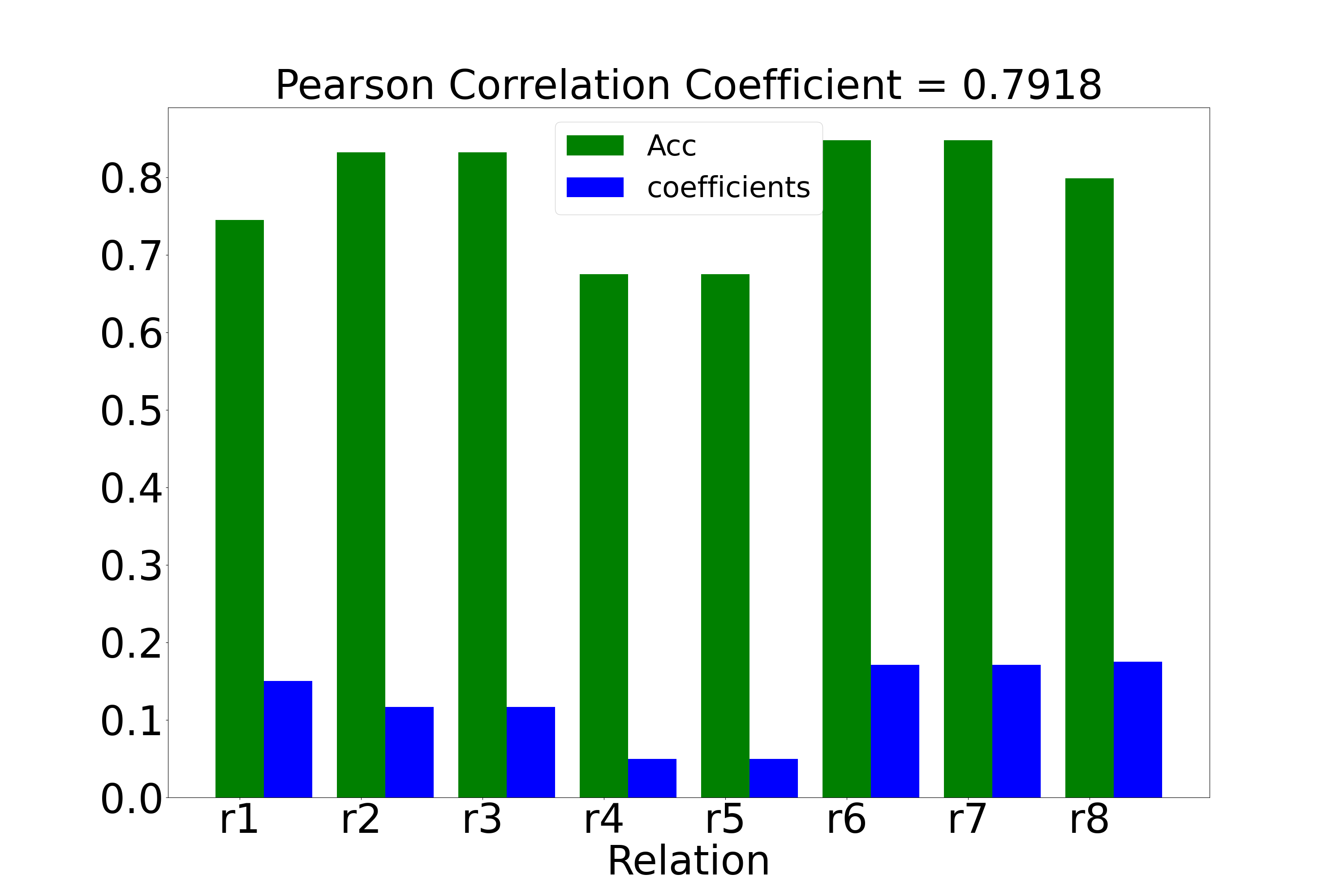}
    \caption{Accuracy  under each single relation and corresponding relational coefficient.}
    \label{coe}
\end{figure}

\section{Experiment}

\label{sec:experiment}


\subsection{Experimental Settings}
\paragraph{Datasets.}
The following four real-world heterogeneous datasets in various fields are  utilized and can be divided into two categories: i) the node type and edge type are both heterogeneous (DBLP~~\cite{fu2020magnn}, ACM~\cite{lv2021we}).
ii) the node type is homogeneous but the edge type is heterogeneous (MUTAG~\cite{schlichtkrull2018modeling}, BGS~\cite{schlichtkrull2018modeling}).
The statistics of the datasets can be found in Table \ref{sta}.
The basic information about datasets is summarized in appendix B.1.

\paragraph{Baselines.}
To test the performance of the proposed EMR-GNN,  
we compare it with five  state-of-the-art baselines.
Among them, GCN~\cite{kipf2016semi} and GAT~\cite{velivckovic2017graph} as two popular approaches are included.
In addition, we compare with the heterogeneous graph model HAN~\cite{wang2019heterogeneous}, since HAN can also employ multiple relations.
Two models that are specially designed for multi-relational graphs are compared, i.e., RGCN~\cite{schlichtkrull2018modeling}  and e-RGCN~\cite{thanapalasingam2021relational}.


\paragraph{Parameter settings.}
We implement EMR-GNN based on Pytorch.\footnote{\url{https://pytorch.org/}}
For $f\left(\mathbf{X};\mathbf{W}\right)$  and $g_{\theta}(\cdot)$,
we choose one layer MLP for DBLP and ACM, and linear layers for MUTAG and BGS. 
We conduct 10 runs on all datasets with the fixed training/validation/test split for all experiments.
More implementation details can be seen in appendix B.3.


\subsection{Node Classification Results}
Table \ref{total} summarizes the performances of EMR-GNN and several  baselines on semi-supervised node classification task.
Since HAN's code uses the heterogeneity of nodes to design meta-paths, we do not reproduce the results of HAN on homogeneous dataset (MUTAG, BGS) with only one type of nodes.
We use accuracy (Acc) and recall metrics for evaluation, and report the  mean and standard deviation of classification accuracy and recall.
We have the following observations:
(1) Compared with all baselines, the proposed EMR-GNN generally achieves the best performance across all datasets on seven of the eight metrics, which demonstrates the effectiveness of our proposed model. 
e-RGCN has a higher recall but a lower accuracy on MUTAG, which may be caused by  overfitting.
(2) Meanwhile, the number of parameters of our model and other baselines are shown in Table \ref{pa}. We can see that 
EMR-GNN  is more parameter efficient than all baselines, 
 i.e., $\mathcal{O}\left(2 d^{2}+K|\mathcal{R}|\right)$,
but achieves maximum relative improvements of $4.14\%$ than RGCN on BGS.
It means that EMR-GNN largely overcomes the over-parameterization in previous multi-relational GNNs.

\subsection{Model Analysis}
\paragraph{Alleviating over-smoothing problem.}
As mentioned before, EMR-GNN is able to alleviate over-smoothing issue.
Here, we take one typical single-relation GCN (GAT) and one representative multi-relational GCN (RGCN) as baselines to test their performance with different propagation depths, where the results are shown in
Figure.\ref{nlayer}. We have the following observations:
(1) Our model significantly alleviates the over-smoothing problem, 
since there is generally no performance degradation when the depth increases.
This benefits from the  adjustable factor $\mathbf{\lambda_{1}}$ in EMR-GNN, which flexible controls the influence of  node feature information.
In contrast, the performance of RGCN and GAT drops seriously  with increasing depth, 
implying that these models suffer from the over-smoothing problem.
(2) RGCN needs huge storage cost, making it difficult to stack multiple layers.
Cuda out of memory occurs when the propagation depth increases, i.e., DBLP for more than 16 layers, and ACM can merely stack  8 layers. This is not available for capturing  long-range dependencies.


 

\begin{table}[!htb]

\renewcommand{\arraystretch}{1.2}

\small 
\begin{tabular}{l|l|rrrr}
\bottomrule 
Datasets &Method &\multicolumn{4}{c}{Size of training set}\\
\hline
 & & $10\%$ & $15\%$& $20\%$&$25\%$\\
\hline
 DBLP &RGCN &  0.5381& 0.6388& 0.7515&0.7721\\

 & EMR-GNN &   0.8768& 0.9109 &  0.9128& 0.9364 \\
 \hline
 ACM &RGCN & 0.7492 &0.8136 &0.8278&0.8344  \\

 & EMR-GNN &  0.8489 & 0.8654 & 0.8739 & 0.8753  \\
\bottomrule 
\end{tabular}
\caption{Classification accuracy w.r.t. different training set.}
\label{size}
\end{table}
\paragraph{Alleviating over-parameterization  problem.}
To further illustrate the advantages of alleviating over-parameterization,
we verify EMR-GNN  with small-scale training samples.
We conduct experiments on EMR-GNN and RGCN using two datasets. We only select a small part of nodes from original training samples as the new training samples. 
As can be seen in Table \ref{size}, EMR-GNN consistently outperforms RGCN with different training sample ratios, which again validates the superiority of the proposed method.
One reason is that  a limited number of parameters in EMR-GNN can be fully trained with few samples.
In contrast, RGCN with excess parameters requires large-scale training samples as the number of relations increases.
The time complexity is analyzed in appendix C.

\noindent
\paragraph{Analysis of relational coefficients.}
Besides the performance, 
we further show that EMR-GNN can produce reasonable relational  coefficients.
To verify the ability of relational coefficients learning, 
taking ACM dataset as example, we evaluate the classification performance under each single relation.
The classification accuracy and the corresponding relational coefficient value are reported in Figure \ref{coe}.
We can see that basically, the relation which achieves better accuracy is associated with a larger coefficient. Moreover, we compute the pearson correlation coefficient between the accuracy of a single relation and its relational coefficient, which is 0.7918, well demonstrating that they are  positively correlated.

\noindent
\paragraph{Visualization.}
For a more intuitive comparison, we conduct the task of visualization on
DBLP dataset.
We plot the output embedding on the last layer of EMR-GNN and three baselines (GCN, GAT and RGCN) using t-SNE~\cite{van2008visualizing}. 
All nodes in Figure \ref{vis} are colored by the ground truth labels.
It can be observed that EMR-GNN performs best, since the significant boundaries between nodes of different colors, and the relatively dense distribution of nodes with the same color.
However, the nodes with different labels of GCN and RGCN are mixed together.
In addition, we also calculate the silhouette coefficients (SC) of the classification results of different models, and EMR-GNN achieves the best score, furthering indicating that the learned representations of EMR-GNN have a clearer structure.

\section{Conclusion}
\label{sec:conclusion}
In this work, 
we study how to design multi-relational graph neural networks from the perspective of optimization objective. We propose an ensemble optimization framework, and derive a novel ensemble message passing layer. Then we present the ensemble multi-relational GNNs (EMR-GNN), which has nice 
relationship with multi-relational/path  personalized PageRank and can recover some popular GNNs.
EMR-GNN not only is designed with clear objective function, but also can well alleviate  over-smoothing and over-parameterization issues.
Extensive experiments demonstrate the superior performance of EMR-GNN over the several state-of-the-arts.

\clearpage

\section*{Acknowledgements}
The research was supported in part by the National Natural Science Foundation of China (Nos. 61802025, 61872836, U1936104) and Meituan.

\begin{thebibliography}{}


\bibitem[\protect\citeauthoryear{Beck and Teboulle}{2003}]{beck2003mirror}
Amir Beck and Marc Teboulle.
\newblock Mirror descent and nonlinear projected subgradient methods for convex
  optimization.
\newblock {\em Operations Research Letters}, 31(3):167--175, 2003.


\bibitem[\protect\citeauthoryear{Chen and Ye}{2011}]{chen2011projection}
Yunmei Chen and Xiaojing Ye.
\newblock Projection onto a simplex.
\newblock {\em arXiv preprint arXiv:1101.6081}, 2011.

\bibitem[\protect\citeauthoryear{Defferrard \bgroup \em et al.\egroup
  }{2016}]{defferrard2016convolutional}
Micha{\"e}l Defferrard, Xavier Bresson, and Pierre Vandergheynst.
\newblock Convolutional neural networks on graphs with fast localized spectral
  filtering.
\newblock {\em Advances in neural information processing systems},
  29:3844--3852, 2016.

\bibitem[\protect\citeauthoryear{Fu \bgroup \em et al.\egroup
  }{2020}]{fu2020magnn}
Xinyu Fu, Jiani Zhang, Ziqiao Meng, et~al.
\newblock Magnn: Metapath aggregated graph neural network for heterogeneous
  graph embedding.
\newblock In {\em Proceedings of The Web Conference 2020}, pages 2331--2341,
  2020.

\bibitem[\protect\citeauthoryear{Geng \bgroup \em et al.\egroup
  }{2012}]{geng2012ensemble}
Bo~Geng, Dacheng Tao, Chao Xu, et~al.
\newblock Ensemble manifold regularization.
\newblock {\em IEEE Transactions on Pattern Analysis and Machine Intelligence},
  34(6):1227--1233, 2012.

\bibitem[\protect\citeauthoryear{Gilmer \bgroup \em et al.\egroup
  }{2017}]{gilmer2017neural}
Justin Gilmer, Samuel~S Schoenholz, Patrick~F Riley, et~al.
\newblock Neural message passing for quantum chemistry.
\newblock In {\em International conference on machine learning}, pages
  1263--1272. PMLR, 2017.


\bibitem[\protect\citeauthoryear{Huang \bgroup \em et al.\egroup
  }{2020a}]{huang2020skipgnn}
Kexin Huang, Cao Xiao, Lucas Glass, et~al.
\newblock Skipgnn: predicting molecular interactions with skip-graph networks.
\newblock {\em Scientific reports}, 10(1):1--16, 2020.


\bibitem[\protect\citeauthoryear{Ji \bgroup \em et al.\egroup
  }{2021a}]{ji2021heterogeneous}
Houye Ji, Xiao Wang, Chuan Shi, et~al.
\newblock Heterogeneous graph propagation network.
\newblock {\em IEEE Transactions on Knowledge and Data Engineering}, 2021.


\bibitem[\protect\citeauthoryear{Kipf and Welling}{2016}]{kipf2016semi}
Thomas~N Kipf and Max Welling.
\newblock Semi-supervised classification with graph convolutional networks.
\newblock {\em arXiv preprint arXiv:1609.02907}, 2016.

\bibitem[\protect\citeauthoryear{Klicpera \bgroup \em et al.\egroup
  }{2018}]{klicpera2018predict}
Johannes Klicpera, Aleksandar Bojchevski, and Stephan G{\"u}nnemann.
\newblock Predict then propagate: Graph neural networks meet personalized
  pagerank.
\newblock {\em arXiv preprint arXiv:1810.05997}, 2018.

\bibitem[\protect\citeauthoryear{Lee \bgroup \em et al.\egroup
  }{2013}]{lee2013pathrank}
Sangkeun Lee, Sungchan Park, Minsuk Kahng, et~al.
\newblock Pathrank: Ranking nodes on a heterogeneous graph for flexible hybrid
  recommender systems.
\newblock {\em Expert Systems with Applications}, 40(2):684--697, 2013.


\bibitem[\protect\citeauthoryear{Liu \bgroup \em et al.\egroup
  }{2021}]{liu2021elastic}
Xiaorui Liu, Wei Jin, Yao Ma, et~al.
\newblock Elastic graph neural networks.
\newblock In {\em International Conference on Machine Learning}, pages
  6837--6849. PMLR, 2021.


\bibitem[\protect\citeauthoryear{Lv \bgroup \em et al.\egroup
  }{2021}]{lv2021we}
Qingsong Lv, Ming Ding, Qiang Liu, et~al.
\newblock Are we really making much progress? revisiting, benchmarking and
  refining heterogeneous graph neural networks.
\newblock In {\em Proceedings of the 27th ACM SIGKDD Conference on Knowledge
  Discovery \& Data Mining}, pages 1150--1160, 2021.

\bibitem[\protect\citeauthoryear{Ma \bgroup \em et al.\egroup
  }{2021}]{ma2021unified}
Yao Ma, Xiaorui Liu, Tong Zhao, et~al.
\newblock A unified view on graph neural networks as graph signal denoising.
\newblock In {\em Proceedings of the 30th ACM International Conference on
  Information \& Knowledge Management}, pages 1202--1211, 2021.


\bibitem[\protect\citeauthoryear{Oono and Suzuki}{2019}]{oono2019graph}
Kenta Oono and Taiji Suzuki.
\newblock Graph neural networks exponentially lose expressive power for node
  classification.
\newblock {\em arXiv preprint arXiv:1905.10947}, 2019.


\bibitem[\protect\citeauthoryear{Schlichtkrull \bgroup \em et al.\egroup
  }{2018}]{schlichtkrull2018modeling}
Michael Schlichtkrull, Thomas~N Kipf, Peter Bloem, et~al.
\newblock Modeling relational data with graph convolutional networks.
\newblock In {\em European semantic web conference}, pages 593--607. Springer,
  2018.


\bibitem[\protect\citeauthoryear{Thanapalasingam \bgroup \em et al.\egroup
  }{2021}]{thanapalasingam2021relational}
Thiviyan Thanapalasingam, Lucas van Berkel, Peter Bloem, and Paul Groth.
\newblock Relational graph convolutional networks: A closer look.
\newblock {\em arXiv preprint arXiv:2107.10015}, 2021.

\bibitem[\protect\citeauthoryear{Van~der Maaten and
  Hinton}{2008}]{van2008visualizing}
Laurens Van~der Maaten and Geoffrey Hinton.
\newblock Visualizing data using t-sne.
\newblock {\em Journal of machine learning research}, 9(11), 2008.

\bibitem[\protect\citeauthoryear{Vashishth \bgroup \em et al.\egroup
  }{2019}]{vashishth2019composition}
Shikhar Vashishth, Soumya Sanyal, Vikram Nitin, and Partha Talukdar.
\newblock Composition-based multi-relational graph convolutional networks.
\newblock {\em arXiv preprint arXiv:1911.03082}, 2019.

\bibitem[\protect\citeauthoryear{Veli{\v{c}}kovi{\'c} \bgroup \em et al.\egroup
  }{2017}]{velivckovic2017graph}
Petar Veli{\v{c}}kovi{\'c}, Guillem Cucurull, Arantxa Casanova, et~al.
\newblock Graph attention networks.
\newblock {\em arXiv preprint arXiv:1710.10903}, 2017.

\bibitem[\protect\citeauthoryear{Wang \bgroup \em et al.\egroup
  }{2019}]{wang2019heterogeneous}
Xiao Wang, Houye Ji, Chuan Shi, et~al.
\newblock Heterogeneous graph attention network.
\newblock In {\em The World Wide Web Conference}, pages 2022--2032, 2019.


\bibitem[\protect\citeauthoryear{Xu \bgroup \em et al.\egroup
  }{2018b}]{xu2018representation}
Keyulu Xu, Chengtao Li, Yonglong Tian, et~al.
\newblock Representation learning on graphs with jumping knowledge networks.
\newblock In {\em International Conference on Machine Learning}, pages
  5453--5462. PMLR, 2018.


\bibitem[\protect\citeauthoryear{Yang \bgroup \em et al.\egroup
  }{2021}]{yang2021graph}
Yongyi Yang, Tang Liu, Yangkun Wang, et~al.
\newblock Graph neural networks inspired by classical iterative algorithms.
\newblock In {\em International Conference on Machine Learning}, pages
  11773--11783. PMLR, 2021.

\bibitem[\protect\citeauthoryear{Yu \bgroup \em et al.\egroup
  }{2021}]{yu2021knowledge}
Donghan Yu, Yiming Yang, Ruohong Zhang, et~al.
\newblock Knowledge embedding based graph convolutional network.
\newblock In {\em Proceedings of the Web Conference 2021}, pages 1619--1628,
  2021.

\bibitem[\protect\citeauthoryear{Yun \bgroup \em et al.\egroup
  }{2019}]{yun2019graph}
Seongjun Yun, Minbyul Jeong, Raehyun Kim, et~al.
\newblock Graph transformer networks.
\newblock {\em Advances in Neural Information Processing Systems},
  32:11983--11993, 2019.


\bibitem[\protect\citeauthoryear{Zhu \bgroup \em et al.\egroup
  }{2021}]{zhu2021interpreting}
Meiqi Zhu, Xiao Wang, Chuan Shi, et~al.
\newblock Interpreting and unifying graph neural networks with an optimization
  framework.
\newblock In {\em Proceedings of the Web Conference 2021}, pages 1215--1226,
  2021.

\end{thebibliography}


\end{sloppypar}
\end{document}